\documentclass[11pt]{article}

\usepackage{standard_note} 

\usepackage{graphicx}
\usepackage{caption}
\usepackage{subcaption}
\usepackage{amssymb}
\usepackage{bbm}
\usepackage[normalem]{ulem}
\usepackage{enumitem}
\usepackage{float}

\usepackage{tikz,pgfplots}
\usepackage{textpos}
\usepackage{pgfplotstable}
\usepackage{array}
\usepackage{booktabs}

\pgfplotsset{compat=1.14}
\usepgfplotslibrary{groupplots}
\usepgfplotslibrary{dateplot}
\usepgfplotslibrary{fillbetween}

\graphicspath{ {./plots/} }

\pgfplotsset{select coords between index/.style 2 args={
    x filter/.code={
        \ifnum\coordindex<#1\fi
        \ifnum\coordindex>#2\fi
    }
}}

\usetikzlibrary{matrix,arrows,decorations.pathmorphing}

\usepackage[
backend=bibtex,
url=false,isbn=false,doi=false,
date=year,
hyperref=auto,
style=alphabetic,
natbib=true,
sorting=nty,
firstinits=true,
maxnames=2, maxbibnames=10,
]{biblatex}

\definecolor{DarkBlue}{rgb}{0,0,0.7} 

\long\def\comment#1{}

\usepackage{dsfont}

\usepackage[vlined,ruled,linesnumbered]{algorithm2e}
\usepackage{authblk}

\title{Image recognition from raw labels collected without annotators}

\bibliography{Machine_Learning.bib} 

\author[*]{Fatih Furkan Yilmaz}
\author[$\dagger$,*]{Reinhard Heckel}

\affil[*]{Dept. of Electrical and Computer Engineering, Rice University}
\affil[$\dagger$]{Dept. of Electrical and Computer Engineering, Technical University of Munich}

\begin{document}
\maketitle

\begin{abstract}
Image classification problems are typically addressed by first collecting examples with candidate labels, 
second cleaning the candidate labels manually, 
and third training a deep neural network on the clean examples. 
The manual labeling step is often the most expensive one as it requires workers to label millions of images.
In this paper we propose to work without any explicitly labeled data by 
i) directly training the deep neural network on the noisy candidate labels, and ii) early stopping the training to avoid overfitting.
With this procedure we exploit an intriguing property of standard overparameterized convolutional neural networks trained with (stochastic) gradient descent: 
Clean labels are fitted faster than noisy ones. 
We consider two classification problems, a subset of ImageNet and CIFAR-10. For both, we construct large candidate datasets without any explicit human annotations, that only contain 10\%-50\% correctly labeled examples per class. We show that training on the candidate examples and regularizing through early stopping gives higher test performance for both problems than when training on the original, clean data. 
This is possible because the candidate datasets contain a huge number of clean examples, and, as we show in this paper, the noise generated through the label collection process is not nearly as adversarial for learning as the noise generated by randomly flipping labels.
\end{abstract}

\section{Introduction}

Much of the recent success in building machine learning systems for image classification can be attributed to training deep neural networks on large, humanly annotated datasets, such as ImageNet~\citep{deng2009imagenet} or CIFAR-10~\citep{torralba_80_2008}. 
However, assembling such datasets is time-consuming and expensive: 
Both ImageNet and CIFAR-10 were constructed by first searching the web for candidate images and second labeling the candidate images by workers to obtain clean labels.
The first step is relatively inexpensive and already yields labeled examples, but the accuracy of those automatically collected, henceforth called candidate labels is low: 
For example, depending on each class, only about 10\%-50\% of the candidate labels we automatically collected from Flickr (as described later) are correct, and only about half of the candidate labels obtained in the process of constructing the CIFAR-10 dataset are correct~\citep{torralba_80_2008}. 
The second step in constructing such a dataset is expensive because it requires either employing expert labelers or asking workers through a crowdsourcing platform for several annotations per image, and aggregating them to a clean label.

\subsection{Contributions}

In this paper we propose to train directly on the noisy candidate examples, effectively skipping the expensive human labeling step. 
To make this work, we exploit an intriguing property of large, overparameterized neural networks: If trained with stochastic gradient descent or variants thereof, neural networks fit clean labels significantly faster than noisy ones. 
This fact is well known, see for example the experiments by~\citet[Fig.~1a]{zhang_understanding_2016}.
What is \emph{not} well known is that this effect is sufficiently strong to enable training on candidate labels only. 
Our idea is that, if neural networks fit clean labels faster than noise, then training them on a set containing clean and wrong labels and early stopping the training resembles training on the clean labels only.

Our main finding is that early stopping the training on candidate examples can enable better image classification performance than when trained on clean labels, provided the total number of clean labels in the candidate training set is sufficiently large. 
We show this on CIFAR-10 and ImageNet, two of the most widely-used image classification benchmarks~\cite{benhamner2017kaggle},
with candidate training sets that we constructed without any human labeling step.
This result questions the expensive practice of building clean humanly labeled training sets, and suggests that it can be better to collect larger, noisier datasets instead.

\paragraph{Better ImageNet performance by learning from candidates:}

We demonstrate that we can achieve state-of-the-art ImageNet classification performance on a subset of the original classes, without learning from the original clean labels.
Specifically, we chose 100 classes for which there are sufficiently many Flickr search results and for which there are no significant semantic overlaps between the keywords and/or the search results.
For those classes, we constructed a new candidate training set by collecting images from Flickr by keyword search, using the keywords of the ImageNet-classes (from the  WordNet hierarchy~\cite{miller1995wordnet}). 
We choose Flickr to collect candidate images because it enables image search unaltered by learning algorithms, unlike what a search engine like Google yields.

Our results show that the performance of ResNet-50~\cite{he2016resnet}, a standard ImageNet benchmark network, improves significantly by training and early stopping on the noisy candidate examples, 
compared to training on the original, high quality ImageNet training set.
Specifically, for the 100-class problem ResNet-50 trained on the original training set achieves a top-1 classification error of 15\%
whereas on our candidate training set, with early stopping, ResNet-50 achieves a better error rate of 10.56\%.

\paragraph{Better CIFAR-10 performance by learning from candidates:}
On CIFAR-10, we achieve higher classification performance by training on candidate labels than any 
standard
model achieves when trained on the original training set. 
The goal of the CIFAR-10 classification problem is to classify 32x32 color images in 10 classes. 
The clean training set consists of 5000 images per class and was obtained by labeling the images from the Tiny Images dataset with expert workers. 
We constructed a new noisy training set consisting of candidate examples only, by picking the images from the Tiny Images dataset with the labels of the CIFAR-10 classes, followed by cleaning to prevent any trivial (same images) or non-trivial (similar images) overlaps with the test set. 
Only about half of the examples in this candidate dataset are correctly labeled. 

We trained 
the best performing networks for CIFAR-10, (ResNet~\citep{he2016resnet}, Shake-Shake~\citep{gastaldi2017shake}, VGG~\citep{simonyan2014vgg}, DenseNet~\citep{huang2017densenet}, PyramidNet~\citep{han2017pyramid} and PNASNet~\citep{liu2018progressive}), and each model achieved significantly higher performance with early stopping than when trained on the original, clean examples.

In numbers, the best performing model trained on the CIFAR-10 training set has 7\% classification error on the CIFAR-10.1 test set~\citep{recht_imagenet_2019},
while the best performing model trained on our candidate training set achieves 5.85\%, with early stopping training of the PyramidNet-110 model, lower than any standard model achieves
when trained on the original training set.

\paragraph{Early stopping is critical:}
Early stopping is critical to achieve the best performance. 
By keeping track of the performance of a large clean subset of the noisy training set, we show that the clean labels are fitted significantly faster than the false labels, and a good point to stop is when most of the clean labels are fitted well, because at this point most of the false labels have not been fitted yet.

\paragraph{Limitations of learning from candidate labels:}
Whether our approach works depends on whether there is significant overlap of the candidate labels; we demonstrate this by showing that when training on ImageNet classes that have a significant semantic overlap, without labeling, yields obviously to high prediction errors for such classes. 

\paragraph{Real label noise is far from adversarial:}
Finally, we study the difference between ``real'' noise in the data obtained through the data collection process (from sources such as search engines and Flickr) and artificially generated noise through randomly flipping the labels of a clean dataset, which is often used in the literature as a proxy for the former. We find that ``real'' noise is more structured and therefore is easier to fit and less harmful to the classification performance in contrast to artificially generated noise, which is harder to fit and more harmful. These findings are consistent with those from the recent paper~\cite{jiang2019synthetic}, which quantifies differences of synthetic and real noise.

\subsection{Related work}

While our work focuses on \emph{exploiting} the fact that large neural networks fit clean labels faster than noise, a large number of related works have shown that neural networks \emph{are} robust to label noise, both in theory and practice~\citep{rolnick2017deep}, and have proposed methods to further robustify the networks~\citep{sukhbaatar2014learning, jindal2016learning}. 
Several other papers have suggested loss function adjustments and re-weighting techniques for noise robust training~\citep{zhang2018generalized, ren2018reweight,tanaka2018joint}.

\citet{guan_who_2017} has demonstrated that the classification performance of deep networks remains initially almost constant when training on partly randomly perturbed MNIST (handwritten digits) training examples. 
A number of recent works have offered explanations for why deep neural networks fit structure in data before fitting noise:
\citet{arpit2017memorization} has shown that loss sensitivity of clean examples is different to noisy examples, 
\citet{ma2018dimensionality} demonstrated that DNNs model low-dimensional subspaces that match the underlying data distribution during the early stages of training but need to increase the dimensionality of the subspaces to fit the noise later in training,
and finally
\citet{li2019gradient,arora_fine-grained_2019} have attributed this in the over-parameterized case to the low-rank structure of the Jacobian at initialization.

There are few works that have exploited the fact that clean examples are fitted faster than noise, with three noteable exceptions:
\citet{Shen_Sanghavi_2019} achieved noise-robustness by proposing a iterative re-training scheme, 
\citet{sun2018limited} proposed systematic early stopping, and
\citet{Song_Ma_Auli_Dauphin_2018} trains networks with large learning rates and uses the resulting loss to separate clean from misslabeled examples.
\citet{li2019gradient} 
provided perhaps the first theoretical justification for early stopping to achieve robustness, by showing that gradient descent with early stopping is provably robust to label noise for overparameterized networks.

Recall that we constructed our own candidate datasets. There are of course other datasets which contain raw noisy labels such as the YFCC100M~\cite{thomee2016yfcc100m} and Webvision~\cite{li2017webvision} datasets, and there are interesting works studying noise-robust learning on those~\cite{li2017learning,jiang2018mentornet, luo2019simple}. 
As mentioned before, we constructed our own candidate dataset to ensure that there are no explicit human labeling steps or algorithms trained on such labels involved in the construction process by i) collecting images from Flickr so that no machine learning is involved in obtaining the candidate images (as there would be with Google image search) and ii) using the original source dataset of the CIFAR-10, Tiny Images, for the CIFAR-10 experiments.
We also note that several others also used Tiny Images as a source for augmenting the clean CIFAR-10 and CIFAR-100 datasets~\citep{carmon2019unlabeled,laine2016temporal} in a semi-supervised learning setting; in contrast we do not use any cleaned labels.

Most of the works on learning from noisy labels have used artificially generated noise on clean, baseline datasets for motivation, experimentation and as  performance metrics. As discussed later in this paper, we find that the performance degradation resulting from overfitting the noisy data is significantly worse on such artificially generated noise, whereas overfitting to the true label noise is not nearly as detrimental. 
The reason is that the true label noise 
generated through the dataset construction process
is highly structured and not adversarial.

Recall that our work exploits the fact that large neural networks fit clean labels faster than noise. An analogous phenomenon is that randomly inititialized and un-trained convolutional image generating networks fit natural images faster than noise, which gives excellent image denoising performance without using any training data~\cite{ulyanov_deep_2018,heckel_deep_2019,heckel_denoising_2020}. In addition to enabling denoising, this property of convolutional neural networks enables regularizing inverse problems such as compressive sensing.

\section{Dataset collection and problem setup}

In this section we describe the construction of our candidate datasets.

\subsection{Construction of a candidate dataset for ImageNet}
In our ImageNet experiments, we  classify a subset of size 100 of the classes, 
which were also used in the selection of the Tiny ImageNet~\cite{wu2017tiny} classes. 
ImageNet is a large dataset of large-scale color images of varying sizes, typically scaled to 256x256. 
The ImageNet dataset was obtained by first collecting candidate examples from Flickr and Google by searching for query keywords from the WordNet hierarchy,
and second obtaining clean labels from the candidate labels by collecting annotations from workers on a crowdsourcing platform.

We constructed a new candidate training set by searching for the keywords of the ImageNet synsets in the user-assigned Flickr tags using the publicly available Flickr API. 
The dataset construction did not involve any explicit human labeling. 
We restricted the search to a 5-year period from 2014-07-12 to 2019-07-12, long after the ImageNet construction period, in order to prevent overlaps of our candidate set and the validation set of the ImageNet that we used to evaluate performance. 

We did not filter or further label any of the images and only resized them to 256x256. 
The resulting candidate dataset consists of 1.45M images across the 100 classes, for which the original ImageNet training set contains 130k images. 
Note that the candidate dataset is extremely noisy, our manual inspections revealed the accuracy of the labels to be between 10\%-50\% and to be highly dependent on the specific class and its corresponding search keywords.

We intentionally collected images with Flickr and not with Google or other image search engines as they utilize machine learning algorithms which are trained on clean-labeled examples, and thus involve explicit human labeling steps. 
Moreover, the results of a Google image search change over time. Using Google image search would therefore introduce significant bias to our candidate dataset and would hinder the reproducibility of our experiments.

\begin{figure}[t]
    \centering
    \includegraphics{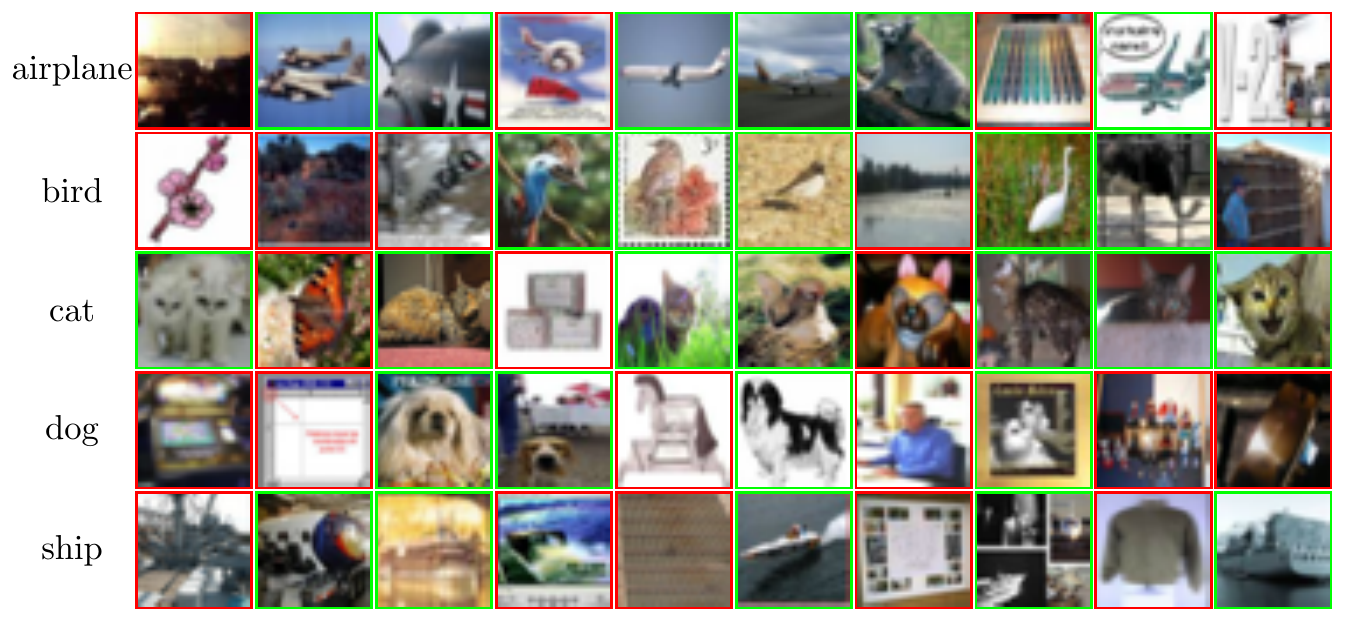}
    \caption{
    \label{fig:random_images}
    Randomly selected images from our noisy training set for some CIFAR-10 classes. Each row corresponds to a class and contains 10 randomly selected images. 
    Note that about half of the images (marked with a red rectangle) are completely unrelated to the class and even to other classes.}
\end{figure}

\subsection{Construction of a candidate dataset for CIFAR-10}

We also constructed a new candidate training set for the CIFAR-10 
classification problem, again without any explicit human labeling effort.
The goal of the CIFAR-10 classification problem is to classify 32x32 color images into the ten classes \{airplane, automobile, bird, cat, deer, dog, frog, horse, ship, truck\}.

The original CIFAR-10 training and test sets as well as our candidate training set are based on the Tiny Images dataset~\citep{torralba_80_2008}.
The Tiny Images dataset consists of 80 million 32x32 images obtained by searching for nearly $80,000$ query keywords or labels in the image search engines Google, Flickr, and others. This yields examples of the form (keyword, image), but the associated labels (keywords) are very noisy, specifically only about 44\% of the labels in the Tiny Images are correct despite various quality controls used by the authors~(see Sec.~III-B in~\citep{torralba_80_2008}), and often an example image is completely unrelated to the label (keyword).

The original, clean CIFAR-10 training and test sets were obtained by first extracting candidates from the noisy examples
of the Tiny Images for each of the ten classes based on the Tiny Images keywords (for example, keywords for the class dog are: puppy, pekingese, chihuahua, etc., see~\citep[Table 6]{recht_imagenet_2019} for a complete list), and then labeling all of the so-obtained images by human experts~\citep{krizhevsky_learning_2009}. This gives a clean training set consisting of 5000 examples per class.

In contrast, we constructed a new and larger (by a factor close to 10) noisy training set by simply taking the subset of the Tiny Images for the respective class according to the same finely grained Tiny Images keywords. This resembles a data collection process without any explicit human labeling.

In our experiments, we measure test performance on the new CIFAR-10.1 test set from~\citet{recht_imagenet_2019}. 
Given that the CIFAR-10.1 test set was also sourced from the Tiny Images, we need to ensure that our candidate training set does not overlap with the CIFAR-10.1 test set in any way. 
Therefore, we carefully removed all test images as well as images that are similar to test images from our noisy training set.
Both is necessary because the extracted subset of the Tiny Images dataset contains many images that are similar to test images up to slight differences in color scale, contrast, original resolution, croppings, and watermarks.
We work with the CIFAR-10.1 test set instead of the original CIFAR-10 test set because it contains fewer images (2000 instead of 10000), and thus made our manual data cleaning process slightly easier. Both datasets share the same distributional properties, and~\citet[Figure 1]{recht_imagenet_2019} found the test performance on both sets to be linearly related.

After removing the similar images, our noisy training set consists of a total of 426.5k images non-uniformly distributed across the 10 classes.
About half of the labels in the noisy training set are wrong, see Figure~\ref{fig:random_images} for an illustration. Also note that they are often completely unrelated and out of distribution---for example a screenshot, a picture of a person, and a wheel are all assigned the label ``dog''.

\section{Training on a large noisy dataset can be  better than training on a clean one}
\label{sec:training_on_noisy}

In this section, we show that training and early stopping standard benchmark deep networks on the large candidate datasets we collected can improve performance over training on the original, clean dataset---despite the many wrong labels---as long as there are sufficiently many clean labels in the noisy dataset. 
Our results show that this is due to neural networks fitting clean labels significantly faster than noisy ones; specifically the clean examples in the training set are fitted during the early stages of training and much faster than the overall training set. Thus, when stopped early, the network is at a state similar to that obtained when only trained on the clean labels in the first place.
This effect enables achieving classification performance by training on very noisy, potentially automatically collected candidate datasets that is on par with the performance obtained by training on very clean, expert annotated datasets.

\begin{figure}[t]
    \centering
    \includegraphics{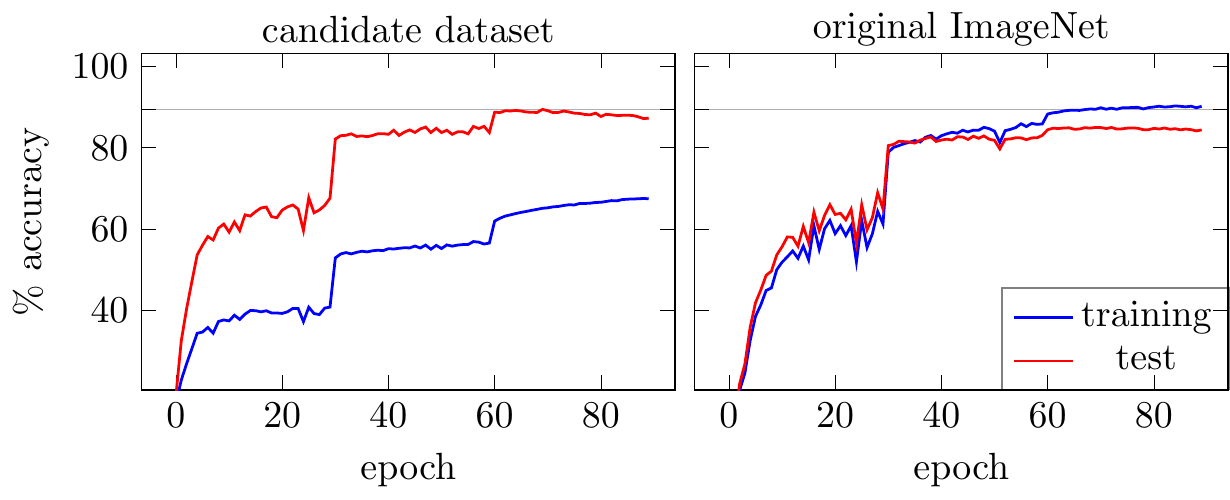}
    \caption{
        \label{fig:imagenet_experiments}
    Classification accuracy on the training and test sets, throughout training with SGD of the ResNet-50 model on our candidate training set (left) that was obtained from Flickr searches and on the original ImageNet training set (right) for classifying 100 ImageNet classes. Training on the candidate images achieves 4.44\% better performance.}
\end{figure}

\begin{figure}[t]
    \centering
    \includegraphics{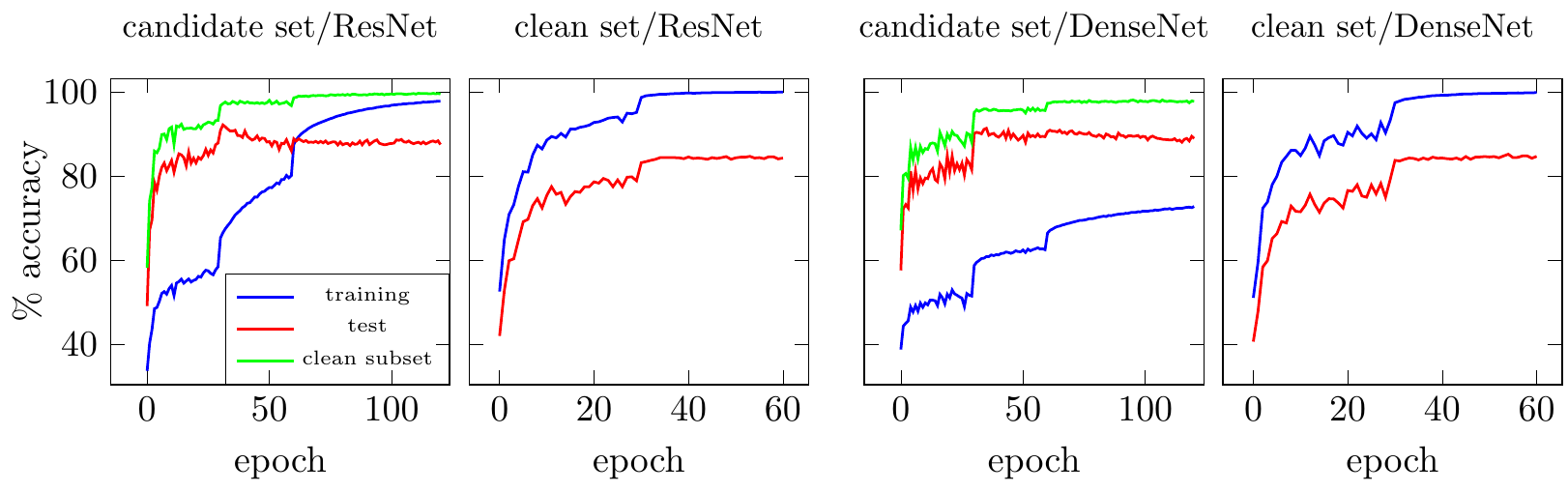}
    \caption{Accuracy on the training set, test set throughout training of the ResNet-18 and DenseNet-110 (k=12) models on our training set (left) and original CIFAR-10 training set (right). For the experiment on our training set, we also show the accuracy on a subset of the training set that is known to have clean labels and corresponds to CIFAR-10 test set.}
    \label{fig:early-stopping-main}
\end{figure}

\subsection{ImageNet experiments}

We first show the training and test accuracy of the standard benchmark ResNet-50 model with the default parameters of the PyTorch library repository~\cite{PyTorch2019} in Figure~\ref{fig:imagenet_experiments}, for training on our candidate training set and for training on the original, clean ImageNet training set (specifically on the subset corresponding to the 100 classes we consider). 
The results show that when trained on our noisy candidate dataset obtained through Flickr searches without explicit human labeling effort, the network achieves significantly better test performance than when trained on the clean dataset (10.56\% vs. 15\% top-1 classification error).
The best performance is achieved when the network is early stopped at about 60 epochs. 
At this optimal early stopping point, essentially all the clean examples but only few of the noisy examples are fitted.

\subsection{CIFAR-10 experiments}

\begin{table}[b!]
    \caption{
        \label{tab:model_comparison}
        Model accuracy (in percent) on the CIFAR-10.1 test set when trained on the original CIFAR-10 training set and our noisy training set. Number of parameters is the total trainable number of parameters of the models and gap is the difference between the corresponding accuracies of the two training sets.}
    \centering

\newcolumntype{C}{>{\centering\arraybackslash}m{4em}}

\pgfplotstableread{
modelName numParameters noisyEarlyAcc noisyConvergeAcc cifarAcc
{PNASNet} 451626 89.55 85.30 81.25
{VGG-16} 14728266 89.9 83.975 84.65
{DenseNet} 769162 91.35 85.52 85.25
{ResNet-18} 11173962 92.15 87.01 85.2
{Shake-Shake} 11914698 93.6 90.22 87.6
{PyramidNet} 28485477 94.15 91.2 87.5
}\modeltable

\pgfplotstablesort[sort key={cifarAcc}]{\sorted}{\modeltable}

\pgfplotstableset{
    sort, sort key={cifarAcc}, sort cmp=float >,
    create on use/gap/.style=
        {create col/expr={\thisrow{noisyEarlyAcc}-\thisrow{cifarAcc}}}}

\pgfplotstabletypeset[
            every head row/.style={
                after row=\midrule},
            every last row/.style={after row=\bottomrule},
            columns/{modelName}/.style={string type,column name=},
            columns/{numParameters}/.style={sci, sci zerofill, sci sep align, precision=2, sci 10e, column name={\#Param}},
            columns/{cifarAcc}/.style={column name={Clean training accuracy}, column type=C},
            columns/{noisyEarlyAcc}/.style={column name={Noisy training accuracy}, column type=C},
            columns/{gap}/.style={column name={Gap}},
            columns={modelName,numParameters,cifarAcc,noisyEarlyAcc, gap},
            ]\modeltable

\end{table}

Similar as for ImageNet, for the CIFAR-10 classification problem, we also achieve significantly better performance when training on our candidate dataset for CIFAR-10. 
Figure~\ref{fig:early-stopping-main} depicts the training and test accuracy of training with stochactic gradient descent with the default PyTorch parameters on our candidate dataset as well as on the clean CIFAR-10 training dataset for ResNet-18.
Again, the results show that when trained on the noisy dataset, with early stopping at about 30 epochs, the network achieves significantly better test performance than when trained on the original clean training set.

Figure~\ref{fig:early-stopping-main} demonstrates our claim that at the optimal early stopping point, almost all of the clean examples, but only very few of the noisy examples are fitted.
This follows by noting that at about 30 epochs, the network achieves almost 100\% training accuracy on a clean subset of the noisy dataset, but only about 65\% overall training accuracy at a false-label-rate of about 50\%.
As mentioned before, for testing we use the CIFAR-10.1 test set, and the clean subset of the training examples is the original CIFAR-10 test set, which is by construction a subset of the candidate training set. 

In practice, the optimal stopping time can be obtained in a data driven way by monitoring the test performance on a small, clean subset of the data, and stopping once close to 100\% training accuracy is achieved on the clean subset of the data.

ResNet is sufficiently overparameterized to perfectly fit the training data, even the mislabeled examples. However, even a network with fewer parameters, such as the DenseNet-110 (k=12) fits the clean data before the network reaches its capacity to fit the noisy labels and the training accuracy plateaus at about 80\%, see Figure~\ref{fig:early-stopping-main}.

We observe this effect across a large number of standard and state-of-the-art models.
In Table~\ref{tab:model_comparison}, 
we provide the main results for the models we have tested on both the original clean CIFAR-10 training set and our candidate training set. The reported accuracies correspond to the best ones achieved during training for each setup. We note that there is a significant improvement in test performance across all models. 
We also observe that the number of parameters does not seem to be directly related to the performance increase of the models.


\section{How clean do the candidate labels have to be?}

In the previous section we found that training on the noisy data gives significantly better performance than training on the clean data. 
Next we demonstrate that this finding relies on the total number of clean labels that are available---if that number grows sufficiently fast with the noise level, then performance becomes increasingly better when training on the noisy training set.

We perform the following experiment to measure performance on increasingly larger and noisier datasets.
We start with the original CIFAR-10 training set and gradually replace a fraction of it with $r=1,3,10$ times more images from the Tiny Images dataset. For example, if the fraction is $0.5$ and $r=3$, that means that the new, noisy dataset consists of 25.000 images of the original clean CIFAR-10 training set and of 75.000 noisy images from our noisy training set. 
Increasing the fraction for some fixed value of $r>1$ thus increases both the dataset size and the number of false labels (say the noise level). 

\begin{figure}[t!]
\centering
\includegraphics{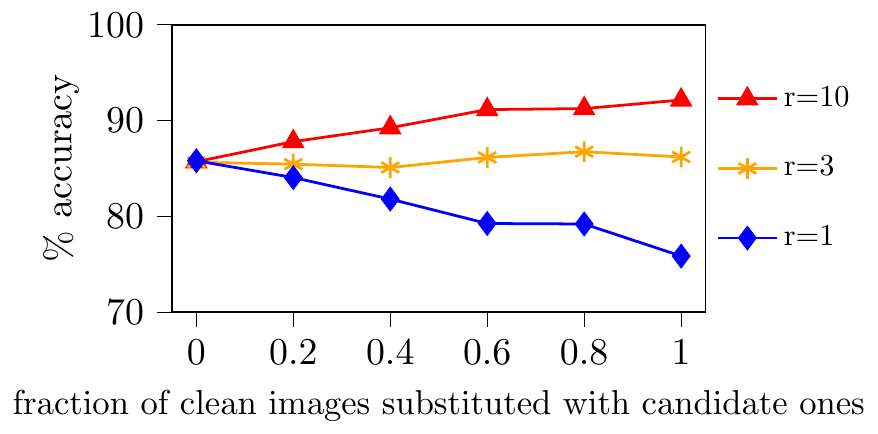}
\caption{\label{fig:CIFAR-10-fraction}
Average test performance
of ResNet-18 for training on datasets with varying number of examples and noise level:
The fraction 0 corresponds to a dataset entirely consisting of clean labels of the CIFAR-10 traing  set; at fraction 0.5, half of the clean images are substituted each with $r$ many images from the noisy dataset.
The results show that that if the number of clean examples grows sufficiently fast with the noise level, the test accuracy improves.
}
\end{figure}

Figure~\ref{fig:CIFAR-10-fraction} depicts the results.
For $r=1$, the total dataset size remains constant but becomes increasingly noisier. Thus the effective number of clean examples in the noisy dataset becomes smaller as the fraction increases. Non-surprisingly, we observe that the test accuracy drops as the noise increases. 
The curve for $r=10$ is more interesting: If we substitute one clean label with $10$ noisy ones, then the test performance increases even though the noise level increases, because the total number of clean examples grows sufficiently fast.

\section{Semantic overlaps and scaling to larger numbers of classes}

\begin{figure}[t]
    \centering
    \begin{minipage}{.4\textwidth}
    \centering
        \includegraphics{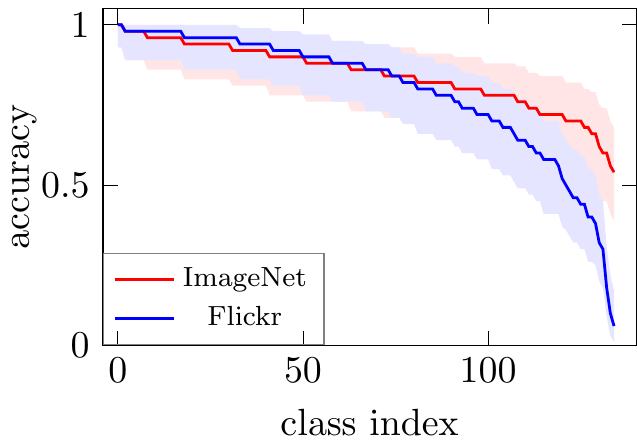}
        \vspace{0.5em}
        \caption{
            \label{fig:flickr_error_distribution}
        Test performance per class, with 95\% Clopper-Pearson confidence intervals, when the model is trained on the 135-class subset of the Flickr with semantic overlaps and the original ImageNet images with no overlaps.}
    \end{minipage}%
    \hspace{1cm}%
    \begin{minipage}{.5\textwidth}
        \centering
        \includegraphics{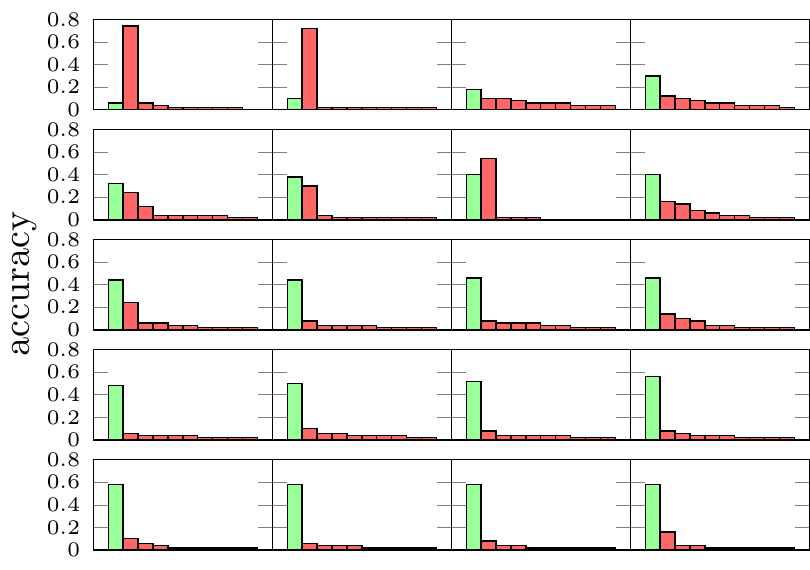}
        \caption{
            \label{fig:conf_bar_plots}
      	Example ImageNet classes (by WordNet IDs) that suffer from test performance degradation when the model is trained on the automatically collected Flickr images due to semantic overlaps in the search keywords (synsets) or search results. Each graph shows the fraction of images classified into the correct class (first bar, in green) and the rest of the other classes in descending order (in red).}
    \end{minipage}%
\end{figure}

Recall that we selected the 100 classes in our ImageNet experiment so that there is no semantic overlap between the classes. Semantic overlap means that the underlying objects are closely related.
Since candidate image collection generally relies on keyword searches, such semantic overlaps can arise if the keywords of classes are semantically closely related, such as ice lolly and ice cream, which are keywords that actually correspond to different ImageNet classes. 
Semantic overlaps result in high in-distribution noise, which is harmful to the classification performance as demonstrated here.
Specifically, as we show in this section, if there are such overlaps then learning from the candidate labels becomes impossible for such classes because we collect images containing the same objects for those classes. 

Semantic overlaps are the main difficulty of learning from candidate images on a large scale, because the larger the number of classes, the more fine-grained they become and thus the more likely such overlaps occur. 
Therefore, for scaling to larger numbers of classes, it is critical to ensure that the objects and keywords of different classes are semantically separated.

We next demonstrate the effects of semantic overlaps on a \emph{larger} subset of ImageNet than our previous experiment, consisting of the previous 100-class problem with an additional 35 classes. Out of the additional 35 classes, some contain semantic overlaps with the original classes. 
Training ResNet-50 on that 135-class candidate dataset and early stopping optimally gives worse classification performance than when training on the corresponding clean ImageNet subset. 
Figure~\ref{fig:flickr_error_distribution}, which depicts the error per class explains why: When trained on the clean ImageNet examples, ResNet-50 has classification performance of more than $60\%$ for all classes, while when trained on the 135-class candidate dataset, it performs better for most classes but extremely poor for others; in particular for those that have semantic overlap.

The latter point is illustrated by Figure~\ref{fig:conf_bar_plots} which shows the distribution of model predictions of the test images of such classes across the correct class as well as the rest of the classes (with non-zero values) as ordered by their contribution to the confusion.
The results show that the primary source of prediction errors are the classes that have semantic overlaps with the actual class. 
For example, the examples of the class `n07615774' with keywords `Ice lolly, lolly, lollipop, popsicle' primarily gets confused with the class `n07614500' with keywords `Ice cream, icecream'.


\section{True versus synthetic label noise}

As mentioned in the introduction, a large number of papers has shown that neural networks fit \emph{randomly} corrupted labels slower than clean ones and/or that neural network are robust to label noise~\citep{zhang_understanding_2016,rolnick2017deep,guan_who_2017, ma2018dimensionality,li2019gradient}. 
The experiments in those works are exclusively for noisy labels generated by randomly flipping the clean labels, which can be considered as in-distribution noise. As we see from Figure~\ref{fig:random_images}, the `true' label noise generated through the image collection process is very different in that it contains lots of out-of-distribution images as well as structured noise. For example,
the keyword `Mouser' used to query images for the cat class by CIFAR-10 yields mostly toy guns and listing images of an electronics distribution company and only a few cat related images. 
Similarly, the keyword `Tomcat' yields only few cat related images as well. Most of the images are related to Java server implementations, the F-14 jet nicknamed Tomcat, and a species of insects.

\begin{figure}[t]
    \centering
    \includegraphics{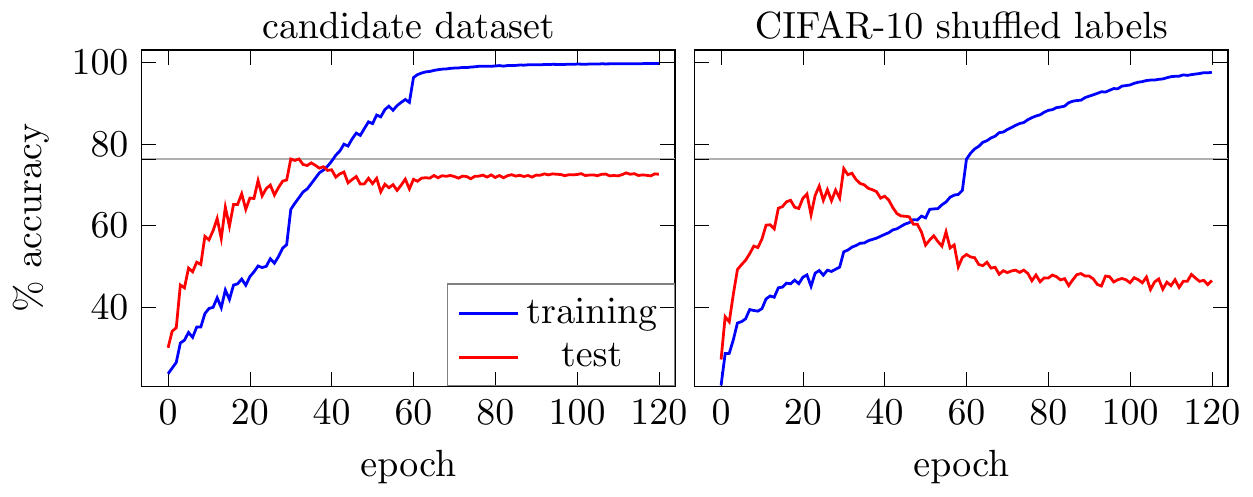}
    \caption{
        \label{fig:distribution_noisy}
    Accuracy on the training set and test set  throughout training of the ResNet-18 model on a $50,000$ image subset of our training set (left) and the same sized original CIFAR-10 training set with the labels permuted such that each class retains $55\%$ clean labels (right).}
\end{figure}

Such structured, ``real'' noise is easier to fit and less harmful to classification performance. 
To demonstrate this, we take the original CIFAR-10 training set and perturb 45\% of the labels by uniformly flipping them, so that this noisy dataset has a similar number of false labels with our candidate dataset. 
We then train on the so-obtained artificially perturbed dataset and a subset of our noisy dataset consisting of the same number of examples (5000 per class). 
This process ensures that the label error probability is approximately the same, but the distribution of the errors is different. 
The results in Figure~\ref{fig:distribution_noisy} show that the data from our noisy dataset is fitted significantly faster because the noise is more structured. 
Moreover, the results show that early stopping is important for both setups, but training until overfitting is not nearly as detrimental to the test performance for training on the candidate dataset than on the permuted one.


\section{Regularization with early stopping: Appealing to linear models}

\newcommand\loss{\mathcal L}
\newcommand\losslin{\mathcal L_{\mathrm{linear}}}
\newcommand\jac{\mathcal J}

For understanding why highly overparameterized neural networks fit structure faster than noise, it is helpful to appeal to linear models.
Specifically, a number of recent works show that highly overparameterized models behave like an associated linear model around the initialization, which in turn offers an explanation why neural networks fit structure faster than no structure~\citep{li2019gradient,du_gradient_2019,du_gradient_2018,arora_fine-grained_2019}.
While the regime where those results hold is extremely overparameterized and thus might not be the regime in which those models actually operate~\citep{Chizat_Oyallon_Bach_2018}, it can provide intuition on the dynamics of fitting clean versus noisy data. 

In this section, we first briefly comment on those recent theoretical results and explain how they provide intuition on why deep networks fit structured data faster than noise, 
and second, inspired by those explanations for linear models, we demonstrate that early stopping the training of non-deep overparameterized \emph{linear} models enables learning from noisy examples as well; thus our findings from the previous section are not unique to deep networks.

\subsection{Existing theory for overparameterized models}

Consider a binary classification problem, and let $f_{\mW} \colon \reals^d \to \reals$ be a neural network with weights $\mW \in \reals^N$, mapping a $d$-dimensional features space to a label. The output label associated with example $\vx$ is interpreted as belonging to class $1$ if $f_{\mW}(\vx)$ is positive and as belonging to class $-1$ if $f_{\mW}(\vx)$ is negative.
Let $\{(y_1,\vx_1),\ldots,(y_n,\vx_n)\}$ be a set of $n$ many training examples. The network is trained by minimizing the empirical loss
$
\loss(\mW) 
=
\frac{1}{2}
\norm[2]{\vy - \vf_{\mW}(\mX)}^2,
$
with gradient descent, where $\vy = [y_1,\ldots,y_n]$ are the labels of the examples and $\vf_{\mW}(\mX) = [f_{\mW}( \vx_1),\ldots,f_{\mW}(\vx_n)]$ are the predictions of the model. 
Let $\jac(\mW) \in \reals^{n \times N}$ be the Jacobian of the mapping from weights to outputs $\vf \colon \reals^{N} \to \reals^n$, and let 
$\mJ \in \reals^{n \times N}$ be a specific matrix that obeys $\mJ \transp{\mJ} = \EX{ \jac(\mW_0) \transp{\jac}(\mW_0)}$, where expectation is with respect to the random initialization~$\mW_0$. 
For two-layer networks, a number of recent works~\citep{li2019gradient,du_gradient_2019,du_gradient_2018,arora_fine-grained_2019} have shown that gradient descent in close proximity around the initialization is well approximated by gradient descent applied to the associated linear loss
\[
\losslin(\mW) = \frac{1}{2} \norm[2]{\vf_{\mW_0}(\mX) + \mJ (\mW - \mW_0) - \vy }^2.
\]
From this, it follows that (see \citep{li2019gradient,arora_fine-grained_2019})
that the prediction error after $t$ iteration of gradient descent with stepsize $\eta$ (provided that $\eta t$ is not too large, and more importantly, provided that the network is sufficiently overparameterized) is given by
\begin{align}
\label{eq:fitcomp}
\norm[2]{\vy - \vf_{\mW_t}(\mX)}^2
\approx
\sum_{i=1}^n (1 - \eta \sigma_i^2)^{2t} (\innerprod{\vy}{\vv_i})^2,
\end{align}
where $\vv_i$ are the eigenvectors and $\sigma_i^2$ are the eigenvalues of the matrix $\mJ \transp{\mJ}$.
Equation~\eqref{eq:fitcomp} shows that the directions associated with large eigenvectors are fitted significantly faster by gradient descent than directions associated with small eigenvalues.
Thus, label vectors $\vy$ that are well aligned with eigenvectors corresponding to large eigenvalues are fitted significantly faster than those corresponding to small eigenvalues. 
The eigenvalues and eigenvectors are determined by the structure of the network. 
It turns out that for neural network the structure is such that the projection of correctly labeled data (i.e., $
\innerprod{\vv_i}{\vy}$, for $\vy$ a set of correct labels) is much more aligned with directions associated with large eigenvalues of $\mJ \transp{\mJ}$, than the  projection of falsely labeled data (i.e., $
\innerprod{\vv_i}{\vy}$, for $\vy$ a set of falsely labeled examples).

This suggests that for highly linear overparameterized models we expect to see similar dynamics, in that for certain models, structure is fitted faster than noise---a property that we demonstrate in the next section.

\subsection{Overparameterized linear models fit clean labels faster than noise}

Finally, we demonstrate that even linear overparameterized models fit clean labels faster than noisy ones, and thus early stopping enables learning from (noisy) candidate examples beyond deep networks.
We consider a random feature model fitted with gradient descent and early stopping. 
The reason for considering a random feature model is that such models can reach performances that are reasonably close to that of a neural network if the number of random features, and thus the size of the model, is sufficiently large.

We consider a random-feature model as proposed in~\citet{coates_ng_lee_2011}. 
We first extract for each example random features through a one-layer convolutional network $G$ with $4000$ random filters, each of size $6\times 6$. 
For each image $\vx$, this gives a feature vector $G(\vx)$ of size $m = 72000$. 
Consider a training set with $N$ examples, and let $\mX \in \reals^{N \times m}$ be the feature matrix associated with the training set, and let $\mY \in \reals^{N \times K}$ be the matrix containing the labels. Specifically, $\mY$ has a one in the $k$-th position of the $i$-th row if the $i$-th training example belongs to the $k$-th class, $k \in \{1,\ldots,K\}$. 
In the training phase, we fit a linear model by minimizing the least-squares loss
$
\loss(\mZ) = \norm[F]{\mY - \mX \mZ}^2
$
with gradient descent. Here, the matrix $\mZ \in \reals^{m\times K}$ specifies the model, along with the random featurizer (i.e., the random convolutional model). 
Given a new example $\vx$, the model then assigns the label $\arg \max_{k} G(\vx) \mZ$. 

\begin{figure}[t]
    \centering
    \includegraphics{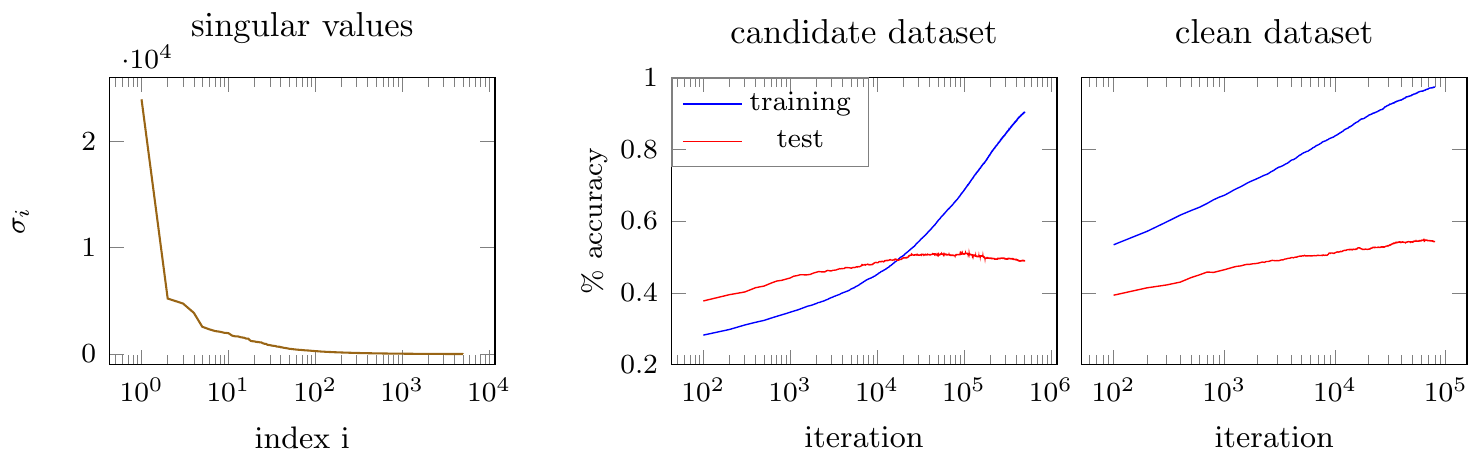}
    \caption{Training a large random feature model on a subset of our candidate training set (middle) and the CIFAR-10 training set (right) with gradient descent. Singular values of the feature matrix (left) shows the low-rank structure for our candidate training set.}
    \label{fig:random_features}
\end{figure}

We trained the model above on two datasets: a clean dataset consisting of 500 examples for each class of the CIFAR-10 training set, and a  candidate dataset consisting of 1500 examples for each class from our candidate dataset.
We measure classification performance as done throughout on the CIFAR-10.1 test set. 
The results, shown in Figure~\ref{fig:random_features}, show that on the clean dataset, this model achieves essentially 100\% training accuracy. On the noisy dataset, similarly as for the deep networks, early stopping improves performance in that the best performance is achieved at about 70,000 iterations of gradient descent, long before the best fit of the model is achieved.

\section*{Code}

Code to reproduce the CIFAR-10 experiments is available at 
\url{https://github.com/MLI-lab/candidate_training}.
The code also contains functions to inspect the training set we constructed and to verify that there is no trivial or non-trivial overlap.

\section*{Acknowledgements}
FFY and RH are partially supported by NSF award IIS-1816986, and acknowledge support of the NVIDIA Corporation in the form of a GPU.
RH would like to thank Ludwig Schmidt for helpful discussions on the CIFAR 10.1 training set.


\printbibliography


\newpage

\appendix
\renewcommand{\thesection}{A\arabic{section}}

\section{\label{app:a-cifar-10}Additional details on the CIFAR-10 experiments}
In this appendix, we provide additional details on the CIFAR-10/Tiny Images experiments.

\subsection{\label{app:data_clenaning}
Avoiding overlap of test and candidate train images
}

In order to make sure that the CIFAR-10.1 test set has no overlap with our noisy dataset, we constructed our noisy dataset by first extracting all images with the respective keywords from the Tiny Images dataset, and then removing all images that are similar to the images in the CIFAR-10.1 test set as follows. 
We first identified the 100 closest images for each individual image in the CIFAR-10.1 test set in both $\ell_2$-distance and Structural Similarity Index (SSIM), and then removed equivalent and similar images after manual visual inspection of these candidates.

\begin{figure}[H]
    \centering
    \includegraphics{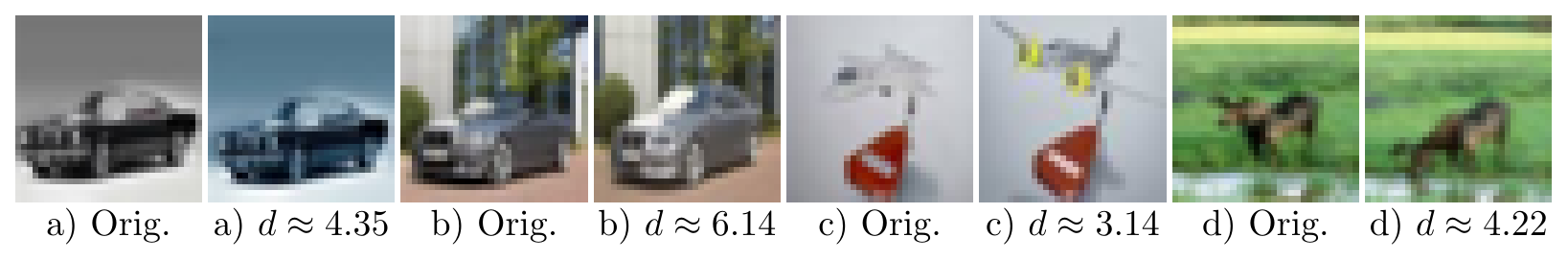}
    \caption{Examples of images that are considered similar for the following reasons: a) Same primary object with different contrasts; b) Different primary objects in the same background c) Same secondary object with different primary objects d) Same object in different poses.}
    \label{fig:sim-main}
\end{figure}

\begin{figure}[H]
\centering
    \includegraphics{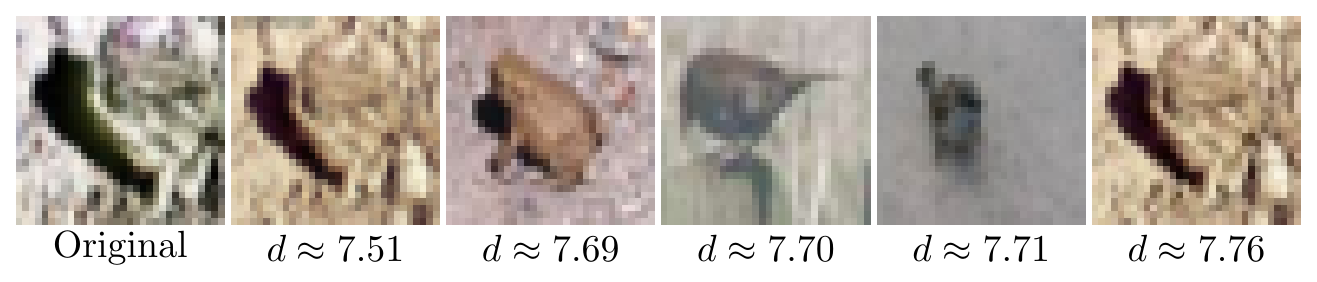}
    \caption{Top-5 closest images to a test image in $\ell_2$-distance (d). 
    The image at distance $d=7.76$ is almost equal to the original one, while a closer one, at $d=7.69$ is a different image.
 }
    \label{fig:sim-cutoff}
\end{figure}

We consider two images similar, if they share the whole or a considerable portion of the same underlying main object, secondary object, or a specific background (see Figure~\ref{fig:sim-main} for examples). 
Performing this step manually is necessary because even images that are relatively far in $\ell_2$-distance or SSIM are often extremely similar, while images relatively close can be significantly different (see Figure~\ref{fig:sim-cutoff} for examples).
We also note that it is too pessimistic to cutoff the images by using a threshold together with a closeness measure because of this wide variety of differences across images that actually contain the same object. Figure~\ref{fig:sim-cutoff} demonstrates an instance of this problem for an image in the `bird' class. There are many other such instances with highly different distance values, which necessitates manual inspection for this part.

\begin{figure}[H]
\centering
    \includegraphics{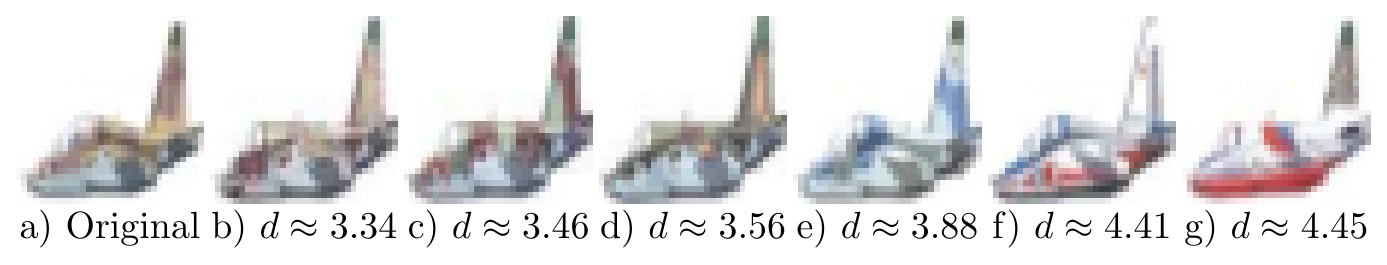}
    \caption{Images that contain the same object in different colors, and are all removed from the training set.}
    \label{fig:sim-94}
\end{figure}

\begin{figure}[H]
\centering
    \includegraphics{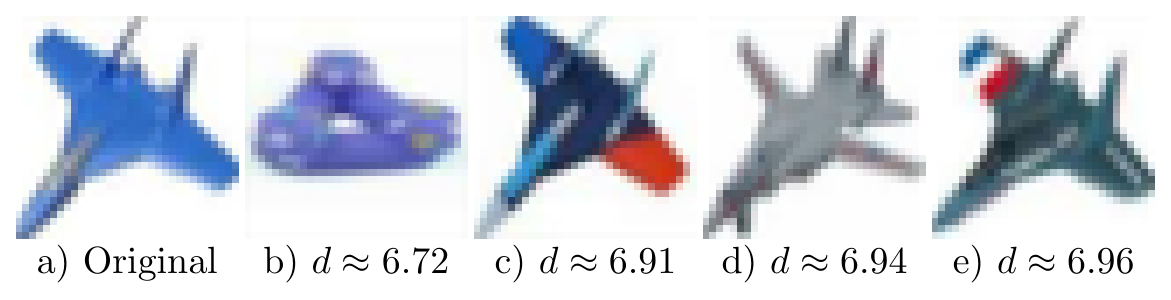}
    \caption{Images that contain the same object in different colors as well as a different object. This demonstrates that images in different colors can have a larger $\ell_2$-distance than completely different images, thus we cannot simply remove images based on $\ell_2$-distance thresholding.}
    \label{fig:sim-146}
\end{figure}

In Figures~\ref{fig:sim-94}--\ref{fig:sim-146}, we show some examples that demonstrate the need for the manual inspection to identify similar images in the test and candidate training sets.
The Figures demonstrate that before the manual filtering step, there are extremely similar images with large $\ell_2$-distance in the test and candidate training datasets. 
The results are similar for when SSIM is used instead of the $\ell_2$-distance.
We note that in addition to the provided similar images, all the test images that are shown here also have identical copies in our training set that can be found without any need for manual labor. This is expected because the CIFAR-10.1 test set is compiled mostly from Tiny Images and it holds true for most of the images in this set.

\newpage
\subsection{\label{app:learningcurves}Learning curves for deep learning models}

\begin{figure}[H]
    \centering
    \includegraphics{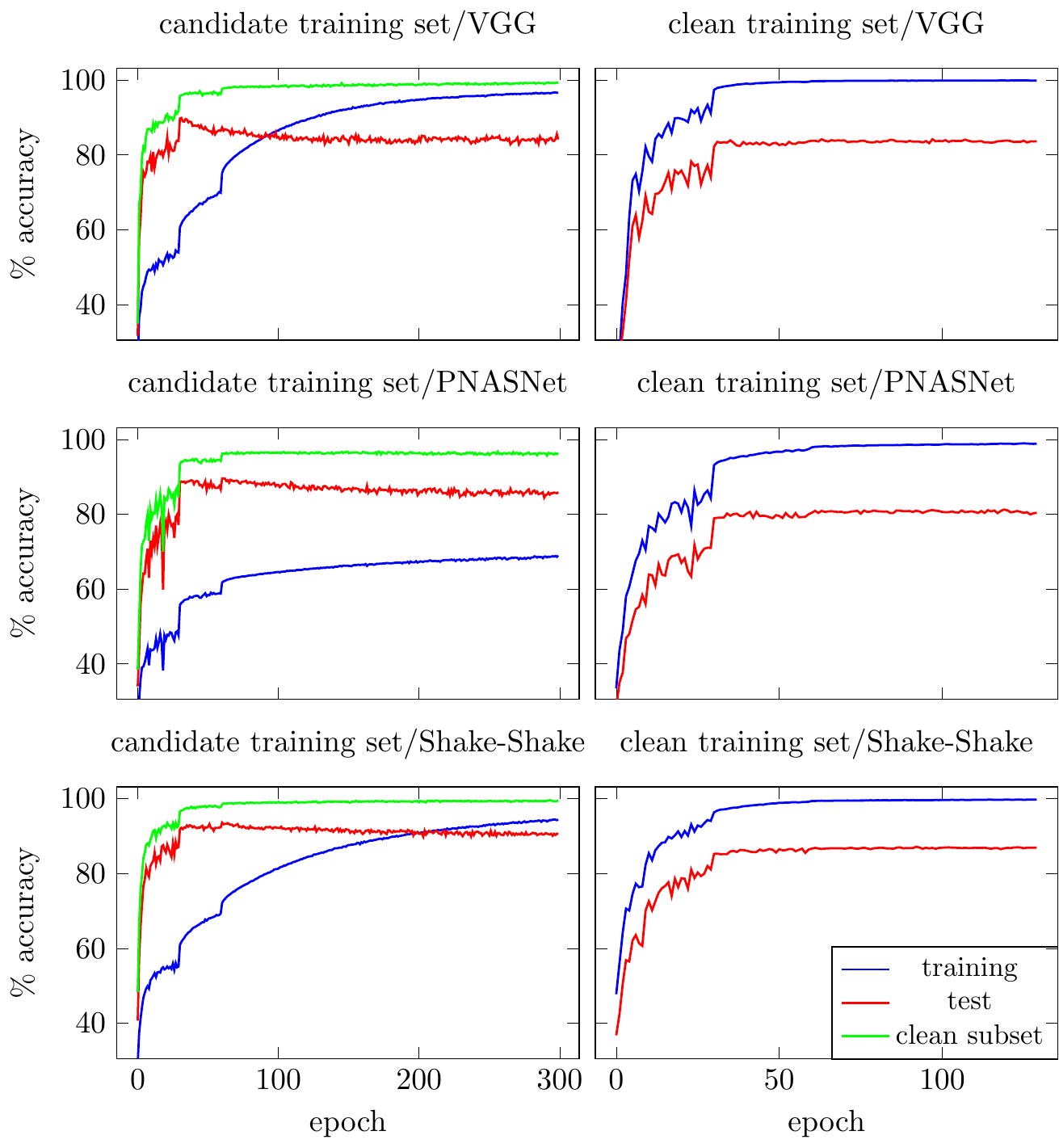}
    \caption{Accuracy on the training set, test set throughout training of the VGG-16, PNASNet, and Shake-Shake-26 2x64d (S-S-I) models on our training set (left) and original CIFAR-10 training set (right). For the experiment on our training set, we also show the accuracy on a subset of the training set that are known to have clean labels and corresponds to CIFAR-10 test set.}
    \label{fig:vgg_group}
\end{figure}

In Figure~\ref{fig:vgg_group}, we display additional learning curves for the models discussed  in the main body of our paper.

\subsection{Error analysis}
\label{app:error_analysis}

\begin{figure}[H]
\begin{center}
\includegraphics{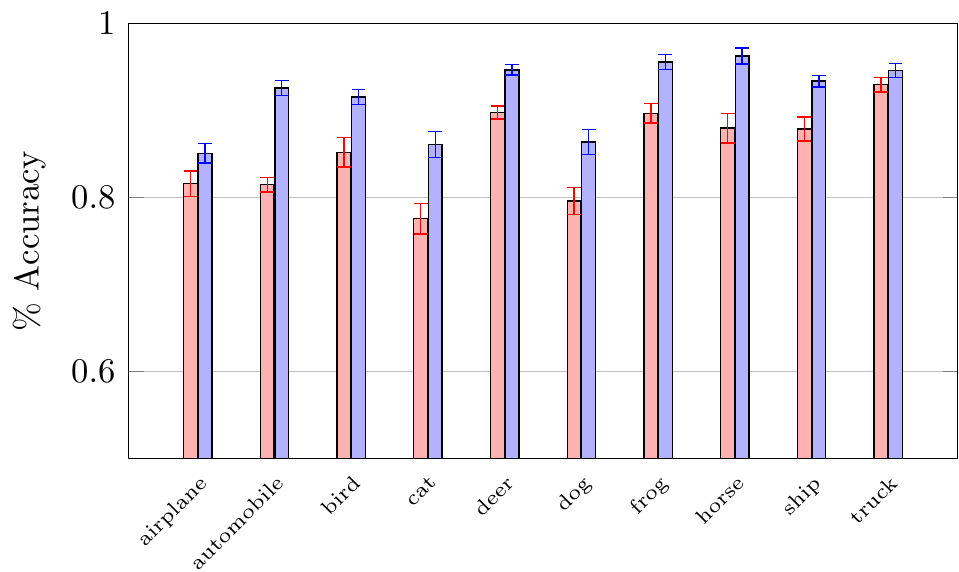}
\end{center}
\caption{
\label{fig:erroranalysis}
Accuracy per class for training ResNet-18 on the original CIFAR-10 training set (red bars) and training on the candidate images with early stopping (blue bars). The plots show that the model trained on the candidate examples has better performance in each class than the model trained on the original training set. The error bars correspond to one standard deviation obtained by averaging over many different initialization.
}
\end{figure}

In this section we take a closer look at the accuracy obtained by training on the candidate dataset. 
Figure~\ref{fig:erroranalysis} shows the accuracy per class for training ResNet-18 on the original CIFAR-10 training set and training on the candidate images with early stopping. The plots show that the model trained on the candidate examples has better performance in each class than the model trained on the original training set. 

\renewcommand{\thesection}{B\arabic{section}}
\setcounter{section}{0}

\section{\label{app:b-imagenet}Additional details on the ImageNet experiments}
In this appendix, we provide additional details  on the ImageNet experiments.

\subsection{Classification of the 135-class subset of the ImageNet with semantic overlaps}
\label{app:class_135_semantic_overlaps}

\begin{figure}[H]
    \centering
    \includegraphics{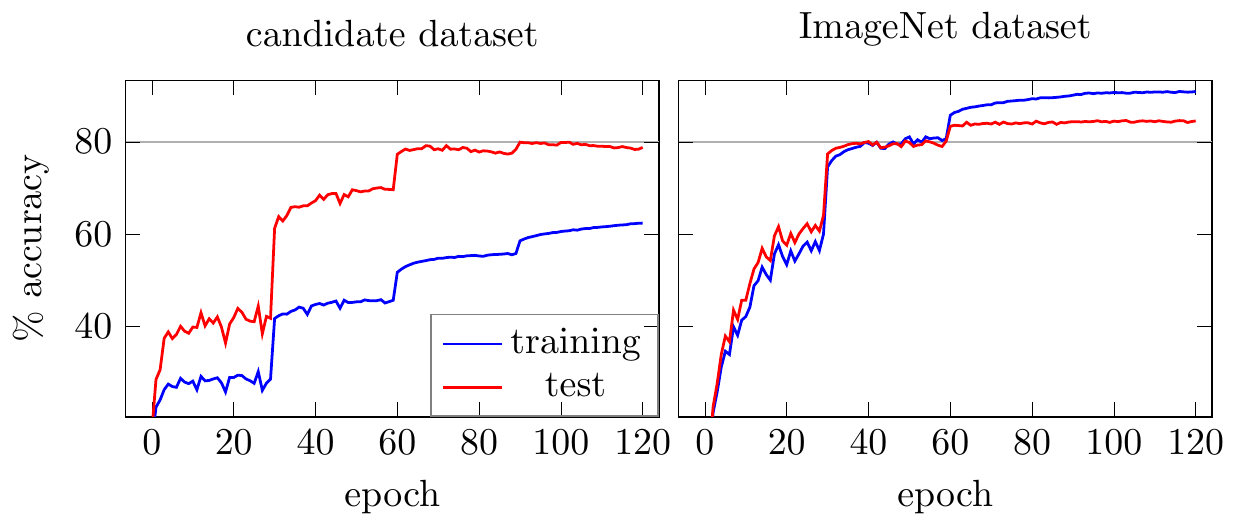}
    \caption{
    Classification accuracy on the training and test sets for the 135-class subset of Flickr with semantic overlaps, throughout training with SGD of the ResNet-50 model on our candidate training set (left) that was obtained from Flickr searches and on the original ImageNet training set (right). Due to degraded test performance of the classes with semantic overlaps, the overall test accuracy when the model is trained on the candidate dataset is notably reduced.}
    \label{fig:flickr_135}
\end{figure}

In this section, we provide the learning curves for training on all the 135 classes discussed in the main body, this includes the original 100 classes along with the 35 classes of which most have semantic overlaps. The results in Figure~\ref{fig:flickr_135} show that the classification performance when training on the 135-class candidate dataset obtained from the Flickr search is worse than when training on the clean ImageNet training set. Note that when training on the 100 classes that do not have significant semantic overlap, the finding was the opposite, i.e., the classifier trained on the candidate examples performed better. The reduced performance is due to the high levels of semantic overlap among the 35 additional classes.

\begin{figure}[H]
    \centering
    \includegraphics{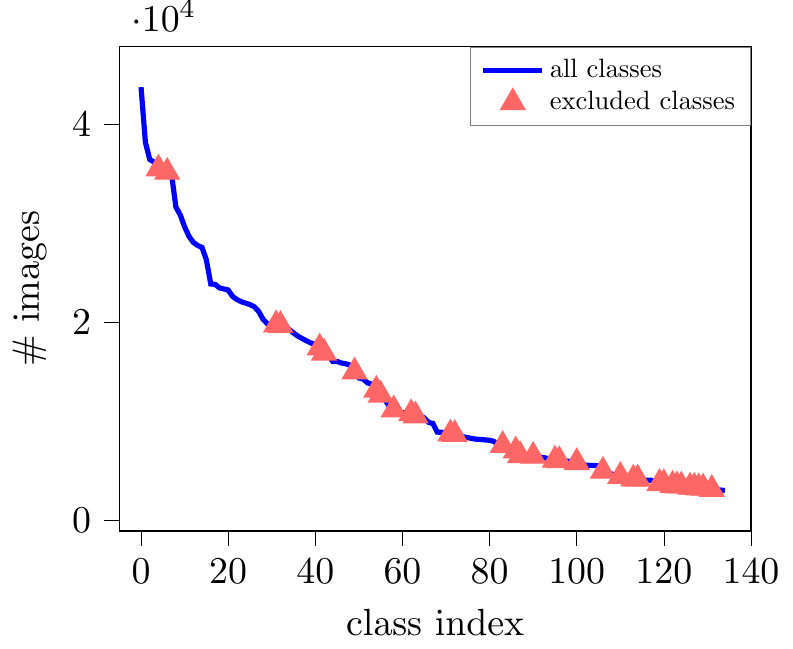}
    \caption{The number of Flickr images collected per class in our candidate training set for the 135 classes, which include the original 100 classes along with the excluded 35 classes. The excluded 35 classes are marked in red. The distribution of these classes in terms of the number of candidate examples indicates that the respective degraded test performance is not directly correlated with the number of training examples.}
    \label{fig:num_imgs}
\end{figure}

We next show that the degraded test performance for the 35 classes with semantic overlaps is not correlated with the number of training examples for these classes. Recall that, we collected our candidate images from Flickr by searching for the keywords of the ImageNet class synsets. Therefore, the number of images among different classes highly varies and depends on the number of search results for the respective keywords. We show the distribution of the training examples of our candidate training set across the 135 classes in Figure~\ref{fig:num_imgs}, where the 35 excluded classes are marked in red. The plots show that the performance of these classes is not directly correlated with the number of training examples obtained from Flickr.

\section{List of the WordNet IDs of the considered classes}
\label{app:wnid_list}

In the following sections, we list the WordNet IDs (WNIDs) of the classes that were considered in our experiments.

\subsection{100-class subset of the ImageNet}
n01914609
n02129165
n04146614
n02509815
n02788148
n02802426
n03388043
n04356056
n04532670
n02486410
n02268853
n04399382
n02843684
n03662601
n01770393
n02950826
n02814533
n01945685
n02132136
n02395406
n01910747
n03670208
n02415577
n03100240
n02165456
n02841315
n01770081
n02279972
n02276258
n02410509
n02793495
n02099601
n03126707
n01944390
n04417672
n02226429
n03976657
n02699494
n01443537
n02906734
n07614500
n02917067
n04540053
n04259630
n07768694
n03804744
n02236044
n04275548
n02837789
n01950731
n07873807
n02481823
n02892201
n03447447
n02106662
n02123045
n07920052
n02058221
n02177972
n04456115
n07753592
n04366367
n02268443
n04099969
n03854065
n07734744
n03160309
n02999410
n02125311
n02423022
n02883205
n02206856
n04008634
n07715103
n01855672
n02190166
n03599486
n02504458
n07749582
n02808440
n01774750
n02769748
n02963159
n04254777
n02948072
n03617480
n02219486
n02317335
n03544143
n03179701
n04465501
n01784675
n04487081
n02256656
n04562935
n02085620
n04507155
n02814860
n03837869
n02480495

\subsection{35-class excluded subset due to semantic overlaps}
n03355925
n02229544
n01984695
n07615774
n02730930
n03201208
n03089624
n03977966
n07579787
n04070727
n04074963
n07747607
n03424325
n06596364
n03770439
n01985128
n04251144
n09193705
n04285008
n03014705
n03584254
n09332890
n09428293
n04398044
n02233338
n02909870
n02977058
n02403003
n02099712
n02791270
n02795169
n03649909
n04067472
n09246464
n01882714
\hspace{4.5em}

\end{document}